\documentclass[runningheads]{llncs}
\usepackage{graphicx}
\usepackage{fancyhdr}
\usepackage{hyperref}

\begin{document}
\title{Compliance Generation for Privacy Documents under GDPR: A Roadmap for Implementing Automation and Machine Learning}
\titlerunning{Compliance Generation for Privacy Documents under GDPR}
%
\author{David Restrepo Amariles\inst{1} \and
Aurore Cl\'ement Troussel\inst{2} \and
Rajaa El Hamdani\inst{3}}
\authorrunning{Restrepo Amariles et al.}
%
\institute{ Associate Professor of Data Law and Artificial Intelligence, HEC Paris. Member of the DATA IA Institute.
\email{restrepo-amariles@hec.fr}\\ \and
Research Assistant, HEC Paris, Smart Law Hub.\\
\email{aurore.troussel@hec.edu} \and
Researcher, Data Science, HEC Paris, Smart Law Hub.\\
\email{el-hamdani@hec.fr}
}
\maketitle              
\begin{abstract}
Most prominent research today addresses compliance with data protection laws through consumer-centric and public-regulatory approaches. We shift this perspective with the Privatech project to focus on corporations and law firms as agents of compliance. To comply with data protection laws, data processors must implement accountability measures to assess and document compliance in relation to both privacy documents and privacy practices. In this paper, we survey, on the one hand, current research on GDPR automation, and on the other hand, the operational challenges corporations face to comply with GDPR, and that may benefit from new forms of automation. We attempt to bridge the gap. We provide a roadmap for compliance assessment and generation by identifying compliance issues, breaking them down into tasks that can be addressed through machine learning and automation, and providing notes about related developments in the Privatech project. 

\end{abstract}
\section{Introduction}
\thispagestyle{fancy}
\cfoot{\small{GDPR Compliance - Theories, Techniques, Tools a Workshop of JURIX 2019. Universidad Polit\'ecnica de Madrid, Madrid, Spain, 11 December 2019}}
Concerns for data protection have become ubiquitous in our increasingly data-driven society. Consumers, citizens, corporations, public organizations, and other social actors are affected by these concerns either as producers or consumers of personal data. Recent regulations, such as the European Union General Data Protection Regulation (GDPR) and the California Consumer Privacy Act (CCPA), have attempted to provide a legal framework to address those concerns by empowering individuals with new rights and by increasing the burden of compliance for firms. Nonetheless, an inconvenient truth lies beneath: a considerable asymmetry of information continues to exist between \textit{data subjects}—i.e., individuals whose personal data is collected and processed, and \textit{data processors}—i.e., those processing and using the data, mainly corporations and governments. This asymmetry of information, together with the inherent technical complexity of the collection, processing, and use of personal data is likely to render ineffective the exercise of individual's rights and the protection of their privacy.

Most prominent research today addresses compliance in relation to data protection laws through consumer-centric or public-regulatory approaches. Initiatives such as Platform Privacy Preferences (P3P), the Usable Privacy Policy Project, the SPECIAL project, Do Not Track (DNT) and CLAUDETTE develop technologies aiming at, on the one hand, empowering consumers to protect their rights, and on the other hand, equipping regulators with tools to tackle non-compliance from data processors. This paper aims to contribute to the current literature on privacy compliance and machine learning by shifting the focus from consumers and regulators to corporations and law firms. Building on existing research, this paper seeks to generate a methodology for scaling the protection of consumers’ rights by automating critical tasks of GDPR compliance in corporations' activities. The adoption of the principle of accountability by recent regulations such as GDPR and CCPA and the shift of the burden of documenting compliance to companies calls for the development of GDPR compliance tools and methods that are fit to corporate settings. However, recent studies on compliance, automation, and machine learning have not yet investigated the actual data practices of corporations and the activity of law firms providing legal advice on compliance with GDPR. We aim to fill this gap through the Privatech project. 

This paper develops a road map to help generate GDPR compliance through automation and machine learning. We assess the extent to which privacy documents of data controllers and processors, and privacy policies, in particular, comply with GDPR. We focus on privacy documents because they are the main instrument to address information asymmetry between data subjects and data processors, and are generally binding. Although GDPR does not mandate privacy policies, they are widely used by firms to communicate data protection information to individuals, regulators and other firms. Additionally, current research shows that companies care about privacy policies because they help to contain risks of data breaches contagion and may contribute to increasing revenue ~\cite{martin2017data}. Moreover, a recent study measuring GDPR’s impact on web privacy conducted among the 500 most visited websites (6,759 websites available in each EU Member State), reveals that 84,5\% have privacy policies, and among these, 50\% were updated following the entry into force of GDPR \cite{degeling2018we}. This study brings to light the need to develop new means for improving the assessment of privacy policies and generating compliance at the source.

\section{From Compliance Consumption to Compliance Generation: Two Case Studies}
Asymmetry of information about how companies collect, process, and use personal data represents a significant risk for both customers and companies. Following on Akerlof \cite{akerlof1978market} seminal work on The Market for Lemons, one may expect that asymmetry of information between companies as data processors, and individuals as data subjects, may lead to a situation in which all companies become bad data processors regarding GDPR requirements. Moreover, a recent study by Kelly D. Martin et al. \cite{martin2017data} underlines the positive correlation between the adoption of good privacy practices and the financial performance of a company (profitability and stability), while a study by Stourm et al. \cite{stourm2019} shows that users’ perception of how companies handle their personal data becomes vital in determining their level of trust and therefore their behavior as consumers. For instance, Jaspreet Bhatia and Travis D. Breaux detected incompleteness and ambiguity in privacy policies and their impact on user’s perception and behavior by using factorial vignette surveys \cite{bhatia2018semantic}. The results reveal a correlation between transparency in data practices and the users’ willingness to share data. More precisely, describing the purposes of the processing increases considerably the willingness of users to share their personal data. Similarly, when the collection of information is done directly upon the users (i.e. not collected from third parties), users’ willingness to share their data also increases \cite{bhatia2018semantic}. 

These studies point to the fact that improving compliance at the source, i.e. in companies acting as data processors, may ultimately be beneficial for consumers. We conducted two case studies to identify the challenges corporations face in complying with GDPR, whether they are addressed internally by their data protection, compliance, and legal departments, or externalized to specialized law firms. 

The first case study consisted in the design of a compliance framework with two senior members of the data protection team of Atos, a leading global information technology and consulting company with a data-intensive activity. Atos is a European leader in providing cloud, cybersecurity, high-performance computing, and e-payment services. Through this case study, we were able to identify some key technological challenges IT companies face to comply with data protection and privacy regulations. 
The second case study consisted in the design of prototype software for the Paris office of the PG/ITC practicing group of one of the top three American global law firms by revenue. This case study identified the tasks performed by external lawyers when providing legal advice to a corporation to ensure GDPR compliance. Through these cases studies, we identified five key challenges:

\begin{enumerate}
    \item Compliance is decentralized: compliance with privacy and data regulation is placed by default on teams and employees leading a project within the company, some of which lack legal knowledge. 
    \item Regulation overload: Operational managers and project leaders within companies face an overload of regulatory information as data, and privacy regulation is expanding quickly. Therefore, they can hardly keep track of new laws and decisions of data protection authorities, let alone of codes of conduct, corporate binding rules, etc.  
    \item Information overload: Information about privacy and data management in contractual relationships is often contained in multiple rather than in a single document. Documents such as contracts, information notices, privacy policies, etc. may contain relevant information regarding the data practices of a contractual party, and it is often difficult for managers to bring this information together and operationalize it in decision-making processes.
    \item Data supply chain: Data processing activities generally imply the participation of multiple actors in the ‘supply chain’. Corporations, whether controllers or processors, may be unable to verify the compliance of privacy and data management policies of subcontractors downstream with their own practices and policies, as well as with those of their clients. 
    \item Offline privacy policies: While a company may have a limited number of online privacy policies, they generally are bound by a large number of internal and contractual documents containing information about privacy and data practices e.g., Contracts in B2B contexts, user notices in B2C. These documents are often neglected in the compliance assessment process.  
\end{enumerate}

\textit{Analysis of Lawyers Tasks for Legal Advice on GDPR Compliance}  
The second case study sought to identify and automate selected tasks performed by lawyers in the law firm, which are complex, mainly due to an overload of information, but which nonetheless are repetitive and with low value. In agreement with the four lawyers who participated in the case study —two partners, one senior associate, and a junior associate— the case study focused on detecting unlawful clauses and missing information in privacy policies.

\textit{Diagnosis} - For reviewing privacy policies, lawyers currently use text editors (i.e., Word office) and share different versions of the privacy policies between them by email. To verify the lawfulness of a privacy policy, lawyers go through the entire text, reading every paragraph, and listing all the information contained within the privacy policy. This is how they detect missing information, which is the most time-consuming task. To assess the lawfulness of a clause, they compared it with an Excel sheet containing the GDPR requirements, clauses previously annulled by a data protection authority or a court, and internal notes from lawyers.

\textit{Tasks} - The issues revealed during the diagnosis phase led us to design a software to automate the detection of keywords and sentences within privacy policy using a dataset of 53 problematic clauses (manually retrieved from privacy policies), 40 problematic words selected by lawyers, and 43 unlawful clauses retrieved from French court decisions and French data protection authority decisions. We used keyword detection for this prototype software to present a basic but operational solution for the firm. As GDPR requires the use of clear and understandable language, the prototype also permitted to obtain a score measuring the readability of paragraphs using the Flesch-Kincaid and Flesch Reading Ease formulas \footnote{\href{https://web.archive.org/web/20160712094308/http:/www.mang.canterbury.ac.nz/writing_guide/writing/flesch.shtml}{Flesch R . How to write plain English.}}. 

\textit{Combining Automation and Machine Learning (ML) to Improve Compliance of Data Processors}
We designed a tool under the name of Privatech, which seeks to assist in the assessment of the compatibility of privacy documents of companies with GDPR, whether this is performed by an internal team of the company or by a law firm. Privatech is designed to assess various legal documents (privacy policies, sub-contractor contracts, binding corporate rules, privacy impact assessment), including those that are internal to data processors and non-accessible for users. At a second stage, we expect to combine this textual approach with an operational assessment of the collection, processing and sharing data practices of companies by analyzing their websites and their data supply chain.

\section{Compliance with GDPR: A Review of Current Methods and Technologies }

This section gives a comprehensive overview of current research and tools related to privacy policies assessment described in the literature.

\subsection{Assessing the clarity and transparency of privacy policies}

Several research studies show good results in assessing the complexity of a privacy policy and subsequent compliance with article 12 of GDPR, requiring the use of plain and unambiguous language.  

\subsubsection{The difficulty to read privacy policies}
As a factor of complexity, research has looked at the difficulty of reading a privacy policy. Researchers applied Flesch Reading Ease and Flesch Kincaid tests to score the readability of privacy policies (Habib and al., 4). The Lexile test developed by Metametrics \cite{litman2019we} and Readable.io are performing tools for assessing the readability of privacy policies \cite{readscoreon}. In addition to these general tests, specific readability metrics have also been developed to analyze privacy policies \cite{anton2004financial}). Indeed, the use of specialized legal vocabulary requires adjustments of general readability tests.

\subsubsection{The vagueness of privacy policies}
The vagueness of privacy policies is a factor of lack of clarity that has been examined through different approaches. 

To detect vagueness of privacy policies, researchers of the CLAUDETTE project combine keywords detection with ML classifiers (Support Vector Machines and bag-of-words). By using this combined approach, they get 81\% of recall of vague clauses with a precision of 32\% for vague clauses \cite{contissa2018claudette}. Lebanoff and Liu \cite{lebanoff2018automatic} investigated both context-aware and context-agnostic models for automatic vagueness detection relying on deep neural network methods (see below on ML classifier). 
Reidenberg et al. \cite{reidenberg2016ambiguity} created a system to compare the relative vagueness of privacy policies in different online sectors. They developed a theory of vague and ambiguous terms, and then used ML and natural language processing (NLP) to automate vagueness assessment. They applied ML methods to retrieve relevant words and expressions related to specific data practices and to annotate privacy policies (see below on classification tasks leveraged by ML). To score the retrieved paragraphs separately, they used NLP tools based on ambiguity’s taxonomy and their own original vagueness scoring method.

\subsubsection{The completeness of privacy policies}
To support transparency, several researchers explored a method to automate the detection of missing information. CLAUDETTE used manual grammars and regular expressions to do so \cite{contissa2018claudette}. Their results showed that some missing mandatory information is more straightforward to detect than others (i.e.the right to lodge a complaint is easier to detect than the description of the purposes of data processing). According to these results, automation of the detection of missing clauses is feasible with an accurate dataset \cite{contissa2018claudette}. Bhatia and Breaux \cite{bhatia2018semantic} also worked on the identification of incompleteness in privacy policies by representing a data practice as a semantic frame. They analyzed five privacy policies and identified 281 data practices from which they identified 17 unique semantic roles they used to assess the completeness of a privacy policy vis-\`a-vis a given data practice (i.e. duration of retention, transfer). 

\subsection{Ensuring data subject empowerment and control on its privacy 
}
Different streams of research have contributed to increase consumers’ control of their data by providing tools to analyze privacy policies. 

\subsubsection{Ensuring the existence of way for users to exercise their rights}
Beyond providing clear and transparent information of users on the processing of their data, a  privacy policy must provide information on the way users could exercise their privacy rights. This requirement has been covered by opt-out choice analysis. Habib and al. \cite{habib2019empirical} analyzed the content of 150 websites, assessing the data deletion option and opt-outs for email communication and targeted advertising. They conclude by offering several suggestions of design and text content that could help users exercising control on their data and privacy. 

\subsubsection{Summarizing privacy policies and highlighting important information}
The summarization could also support the empowerment of users by providing information on the data practices of a data processor. In this respect, Tesfay et al. \cite{bhatia2018semantic} propose a ML tool that presents a risk-based, condensed, and easy to understand summary of privacy policies. We could also mention the work of Wilson and al. \cite{tesfay2018read} and the creation of ASDUS system, which uses a variety of features of text and markup structure to flag titles and prose organization of HTML privacy policies automatically.

\subsubsection{Standardizing privacy best practices}
A more global approach for the protection of data subject consists in designing a methodology for automating the assessment of privacy policies based on data protection and privacy requirements, as developed by the CLAUDETTE project. Researchers created a set of golden standards relying on three types of unlawfulness \cite{gopinath2018supervised} (omission, violation of the prescribed limits of data processing, use of unclear language).

\subsubsection{Modeling GDPR rules}
Modeling GDPR has become a growing research field. Some scholars attempt to model legal reasoning to facilitate compliance with GDPR \cite{palmirani2018legal}. For instance, Bartolini et al. \cite{bartolini2018legal} tried to build executable rules for a computer-assisted compliance assessment by modeling the meaning of GDPR’s articles.

\subsection{Technologies and corpus in the service of privacy compliance assessment}

We identify several methods to make the text of privacy policies machine-readable. These methods could also be applied to other privacy documents.

\subsubsection{Machine learning classifiers}
To better understand the privacy policies, researchers have explored multiple ML classification algorithms. The best results arise from a combination of several classifiers and a keyword approach. 
\begin{enumerate}
    \item Linear regression has been used by Liu et al. \cite{liu2018towards} to identify particularly relevant words for the different categories as classified by other ML classification algorithms.
    \item The Naive Bayes classifier has been used by Zimmeck and Bellovin \cite{zimmeck2014privee}) to identify privacy practices in policy text. This classifier showed strong results in terms of both performance and time-cost on the classification of privacy policy clauses following a privacy risk-based approach and GDPR \cite{bartolini2018legal}.
    \item Lebanoff and Liu \cite{lebanoff2018automatic} explored the auxiliary-classifier generative adversarial networks (AC-GAN) model to discriminate between real/fake privacy policy sentences (depending on their vagueness) and to classify these sentences simultaneously from ‘clear’ to ‘extremely vague’. 
    \item Much research applied support vector machines (SVM) methods to classify sentences of privacy policies. This is the case of CLAUDETTE’s researchers \cite{gopinath2018supervised} who used support vector machines combined with Hidden Markov Models.
    \item Decision tree and Random Forest have been used by Tesfay and al. \cite{bhatia2018semantic}, but they showed lower precision than Naive Bayes and Support Vector Machines.
\end{enumerate}

It seems that the best results arise from a combination of approaches. Wilson and al. \cite{tesfay2018read} used convolutional neural networks (CNNs), logistic regression (LR), and SVMs to classify policy text into privacy practice categories.
Some researchers are relying mainly on Keyword approach \cite{watanabe2015understanding} to analyze privacy policies, while other combines keywords detection and ML classifiers, such as CLAUDETTE \cite{contissa2018claudette}. 

\subsubsection{Annotation methodologies}
The most recent research underlines the need for a combination of crowdsourcing, ML, and NLP to analyze privacy policies at a large scale. The need for human annotation is an obstacle for the development of automated assessment of privacy policies all the more that there is a growing concern in relation to the accuracy of crowdsourced privacy assessments.
Wilson and al. \cite{wilson2016creation} used crowdsourcing privacy policy annotation to create a corpus of website privacy policy analysis and Ammar and al. \cite{ammar2012automatic} also relied on crowdsourcing to categorize data practices within privacy policies. Thus, crowdsourcing is still the traditional way to annotate legal documents. Several research studies aim to define proper human annotation protocols such as Shvartzshnaider and al. \cite{shvartzshnaider2018analyzing}, who presented a method for crowdsourcing contextual integrity annotation on privacy policies. 

The issues arising from human annotation (e.g. costs, interpretation gaps) explain why Wilson and al. \cite{wilson2018analyzing} started from human annotation to define subsequently procedures that require a smaller amount of human labor. They get first encouraging results in semi-automated extraction of relevant information from privacy policies \cite{sadeh2013usable}. Moreover, Wilson and al. \cite{wilson2016creation} recently automated annotation by training a ML algorithm on their OPP-115 corpus (see below). This algorithm annotated over 7000 policies and identified data practices.  
 
\subsubsection{Annotated corpus}
Thanks to the contribution of research projects on privacy policies, annotated corpora are available to improve state of the art and to support the development of tools ensuring data protection.
\begin{enumerate}
    \item The usable privacy policy project’s OPP-115 corpus is composed of a total of 23 194 practice annotations coming from law students using metadata tags on ten data practices such as ‘data retention’ or ‘data security’. 
    \item The APP-350 corpus MAPS is a distributed system for assessing the privacy policies of Apps. In the frame of the usable privacy policy project, Zimmeck and al. \cite{zimmeck2019maps} evaluated a set of more than one million apps. 
    \item The PRIVACYQA is the latest shared corpus of the Usable privacy policy project. It is composed of privacy policy questions and 3500 annotations of relevant answers and aims to support the developing of question-answering tools on privacy \cite{ravichander2019question}. 
    \item The CLAUDETTE project also seeks to create an open and high quality annotated corpus of online privacy policies that follow their golden standards. 
\end{enumerate}

\section{Impacting Data Processors Compliance Strategies: The Privatech Project}

To generate GDPR compliance, data processors must implement the principle of accountability, assessing, and documenting compliance concerning both privacy documents and practices. Privatech is conceived as an agile software, able to host text-assessment features and to extend the scope of its compliance’s assessment to the data processors’ activity. This paper focuses on the legal textual analysis required to enable GDPR compliance. 

Privatech seeks to create a collaborative platform for data processors' privacy experts and lawyers to assist them in the assessment of various legal documents by providing automatic text analysis and suggesting content.

The current analysis of privacy policies relies on a rule-based approach. To strengthen our knowledge engineering system, we build features to gather the users’ feedback and annotations so that our system learns continuously from the users’ legal assessment. Nonetheless, the rules that we have defined until now with legal experts have a high recall but low precision, mainly because some keyword expressions have different meanings when they occur in different contexts. 

Therefore, the next step in the development of Privatech is to build ML algorithms that learn to recognize data practices and GDPR breaches from privacy policies.

In the current state of development of the application, four features have been designed to assess privacy policies: 

\begin{enumerate}
    \item Detection of problematic clauses i.e. segments of text that could breach GDPR. 
    \item Detection of unlawful clauses i.e. unlawful clauses retrieved from decisions of courts or data protection authorities.
    \item Detection of missing information i.e. mandatory information on users’ rights and the processing of personal data.
    \item Assessment of the readability of privacy policies since the GDPR requires the use of plain and clear language.
\end{enumerate}

Each time Privatech flags a clause as problematic or unlawful, the user can see the legal basis of this automatic qualification (either a legal article of GDPR or the reference of a judicial/data protection authority decision). 
Each time a piece of information is listed as missing, the user can see which information is missing and gives feedback about the prediction of the software.

\section{Roadmap for Compliance Generation: Issues and Tasks }

During our preliminary research on the design and development of Privatech, we identified legal and technical issues and subsequent actions to be performed. This section seeks to list all these issue-related tasks to define a scope and strategy for the project. 
\newline

\textbf{Issue 1}: Follow an accurate annotation’s protocol to identify problematic clauses based on a conventional interpretation of the GDPR\\
\textbf{Task}: Identify a list of criteria that a clause/sentence/wording must fulfill to be considered as ‘problematic’ in the sense of potentially incompatible with articles 12, 5 and 6 of the GDPR.\\
\textit{Comments}: To be coherent, the annotation and retrieval of problematic clauses require a common understanding of which kind of clauses/sentences/wording leads to unlawfulness. Primarily, it needs a shared understanding of article 12 requirements on conciseness, transparency, intelligibility, and articles 5,6 on the lawfulness of the processing.
\newline

\textbf{Issue 2}: Use a common terminology to classify clauses depending on their compliance with GDPR.\\
\textbf{Task}: Find a common terminology for the level of compliance of clauses. \\
\textit{Comments}: Privatech defines as ‘problematic’ the clauses/sentences that may be non-compliant with GDPR regarding the current law/case law/guidelines/scholarly work. The clauses that have been considered unlawful by a court or a data protection authority are flagged as ‘unlawful’ on Privatech. Nonetheless, there might be different interpretations of this terminology, as we cannot assure that the same clause will be considered unlawful by another authority or that a new decision will contradict the first one, or even that an appeal or supreme court annuls an existing decision.
\newline

\textbf{Issue 3}: Take into account all the modifications of case law and authorities’ decision practice.\\
\textbf{Task}: Integrate a legal and ruling control of case law and authorities’ decision practice, in order to automate the modification of the database and the parameters of the detection algorithm.\\
\textit{Comments}: A change of case law or authorities’ decision practice or a new decision could invalidate previous assessment of clauses. Thus, Privatech must ensure that such changes are taken into account to modify/delete clauses from its database and that the algorithm will stop flagging clauses newly lawful or will start clauses newly unlawful. 
\newline

\textbf{Issue 4}: Address inconsistency in case law/data protection authorities’ decisions.\\
\textbf{Task}: Define a method to (1) identify inconsistency between decisions and (2) weight decision and their contribution depending on the layer of jurisdiction/normative hierarchy between authorities.
\newline

\textbf{Issue 5}: Address the difference between national laws implementing GDPR within EU Member States\\
\textbf{Task}: Identity, through a comparative legal analysis, differences of interpretation/implementation of GDPR articles within EU Member States. \\
\textit{Comments}: Some articles specify that the Member States have a margin of appreciation toward the implementation of certain requirements. For example, the Member States have such a margin toward the determination of the age limit to accept children’s consent (between 13 and 16 years).
\newline

\textbf{Issue 6}: Reduce the impact of the translation of unlawful clauses on the legal certainty of classification\\
\textbf{Task}: Foresee the potential impact of the translation of unlawful clauses on their legal qualification in another legal order.\\
\textit{Comments}: We retrieve unlawful clauses from French court decisions. We can flag these French language clauses as unlawful as this characterization results from a judge. Can we translate these clauses in English and classify it as unlawful while the translation may change the wording in such a way that the clause could be interpreted differently by an English judge?
\newline

\textbf{Issue 7}: Some clauses may be flagged as problematic, unlawful, and unclear. Some clauses may be problematic/unlawful on the grounds of different legal basis.\\
\textbf{Task}: Identify a method to articulate multiple grounds of detection and multiple legal bases of unlawfulness.
\newline

\textbf{Issue 8}: The semantic analysis of isolated sentences is not sufficient for the determination of their potential unlawfulness/incompleteness.\\
\textbf{Task}: Identify a method that links several sentences and identify relations between them (1), and select the most accurate level of granularity (sentence/segment/paragraphs)(2).\\
\textit{Comments}: some clauses seemed unlawful when isolated because they introduce a section. Thus, we need to take into account the whole privacy policy.
For example, ‘we may collect, among other things, data that are necessary to provide you services, such as:’ and then, the privacy policy lists all the data collected and the related service in different paragraphs. In such a case, the introducing sentence will be wrongly flagged as problematic.
\newline

\textbf{Issue 9}: Recognize important words of a clause to classify it.\\
\textbf{Task}: Annotate clauses which refer to one specific right/information to be provided and extract discriminatory vocabulary.
\newline

\textbf{Issue 10}: Identify clauses related to each data-protection right.\\
\textbf{Task}: Analyze privacy policies to find clauses referring to one particular right.\\
\textit{Comments}: We identified 11 data-protection rights related to GDPR articles such as article 15 about the right of access and article 16 about the right to rectify.
\newline

\textbf{Issue 11}: Adapt general readability scores to GDPR.\\
\textbf{Task}: The main factors that affect the current readability scores are the length of sentences and the difficulty of words. These factors are insufficient to reveal the level of understanding of legal documents that have specific language and syntactic structure. These factors should be taken into account when developing formulas for reading scores.\\
\textit{Comments}: It could be helpful to use a privacy dictionary/ data protection authority notices/ guidelines or even a legal dictionary to enrich the assessment of word difficulty. 
\newline

\textbf{Issue 12}: Use visualization to facilitate the assessment and understanding of privacy policies.\\
\textbf{Task}: Identify the best visualization tools to reflect (1) the readability of privacy policies, (2) the number of problematic clauses crossed with the number of unlawful clauses and missing information (3) the global compliance of a privacy policy compared to other identified privacy policies.\\
\textit{Comments}: the third visualization could rank the concerned privacy policy by referring to the best and worst privacy policies analyzed by our system and the competitors of the company.

\section{Conclusion}
The considerable asymmetry of information that continues to exist between data subjects and data processors could threaten the expected benefits of new privacy regulations such as GDPR. Tools based on automation and ML techniques can help tackle such asymmetry when made available to consumers, regulators, and corporations. This paper laid down the approach of the Privatech project to generate compliance at the source, i.e. in data processors. We expect that the road map put forward with practical issues, individual tasks, and comments on current development by the Privatech project would help the wider privacy community to advance new solutions and tools to protect individual's privacy and enhance data protection rights. 

%
%
%
\bibliographystyle{splncs04}
\bibliography{mybibliography}

\begin{thebibliography}{10}
\providecommand{\url}[1]{\texttt{#1}}
\providecommand{\urlprefix}{URL }
\providecommand{\doi}[1]{https://doi.org/#1}

\bibitem{readscoreon}
Readable: Get your readability score. \url{https://readable.com}, accessed:
  2019-10-30

\bibitem{akerlof1978market}
Akerlof, G.A.: The market for “lemons”: Quality uncertainty and the market
  mechanism. In: Uncertainty in economics, pp. 235--251. Elsevier (1978)

\bibitem{ammar2012automatic}
Ammar, W., Wilson, S., Sadeh, N., Smith, N.A.: Automatic categorization of
  privacy policies: A pilot study. School of Computer Science, Language
  Technology Institute, Technical Report CMU-LTI-12-019  (2012)

\bibitem{anton2004financial}
Anton, A.I., Earp, J.B., He, Q., Stufflebeam, W., Bolchini, D., Jensen, C.:
  Financial privacy policies and the need for standardization. IEEE Security \&
  privacy  \textbf{2}(2),  36--45 (2004)

\bibitem{bartolini2018legal}
Bartolini, C., Lenzini, G., Santos, C.: A legal validation of a formal
  representation of gdpr articles. In: Proceedings of the 2nd JURIX Workshop on
  Technologies for Regulatory Compliance (Terecom) (2018)

\bibitem{bhatia2018semantic}
Bhatia, J., Breaux, T.D.: Semantic incompleteness in privacy policy goals. In:
  2018 IEEE 26th International Requirements Engineering Conference (RE). pp.
  159--169. IEEE (2018)

\bibitem{contissa2018claudette}
Contissa, G., Docter, K., Lagioia, F., Lippi, M., Micklitz, H.W., Pa{\l}ka, P.,
  Sartor, G., Torroni, P.: Claudette meets gdpr: Automating the evaluation of
  privacy policies using artificial intelligence. Available at SSRN 3208596
  (2018)

\bibitem{degeling2018we}
Degeling, M., Utz, C., Lentzsch, C., Hosseini, H., Schaub, F., Holz, T.: We
  value your privacy... now take some cookies: Measuring the gdpr's impact on
  web privacy. arXiv preprint arXiv:1808.05096  (2018)

\bibitem{gopinath2018supervised}
Gopinath, A.A.M., Wilson, S., Sadeh, N.: Supervised and unsupervised methods
  for robust separation of section titles and prose text in web documents. In:
  Proceedings of the 2018 Conference on Empirical Methods in Natural Language
  Processing. pp. 850--855 (2018)

\bibitem{habib2019empirical}
Habib, H., Zou, Y., Jannu, A., Sridhar, N., Swoopes, C., Acquisti, A., Cranor,
  L.F., Sadeh, N., Schaub, F.: An empirical analysis of data deletion and
  opt-out choices on 150 websites. In: Fifteenth Symposium on Usable Privacy
  and Security ($\{$SOUPS$\}$ 2019) (2019)

\bibitem{lebanoff2018automatic}
Lebanoff, L., Liu, F.: Automatic detection of vague words and sentences in
  privacy policies. arXiv preprint arXiv:1808.06219  (2018)

\bibitem{litman2019we}
Litman-Navarro, K.: We read 150 privacy policies. they were an incomprehensible
  disaster. The New York Times  \textbf{12} (2019)

\bibitem{liu2018towards}
Liu, F., Wilson, S., Story, P., Zimmeck, S., Sadeh, N.: Towards automatic
  classification of privacy policy text. School of Computer Science Carnegie
  Mellon University, Pittsburgh, PA, Tech. Rep. CMU-ISR-17-118R and
  CMULTI-17-010  (2018)

\bibitem{martin2017data}
Martin, K.D., Borah, A., Palmatier, R.W.: Data privacy: Effects on customer and
  firm performance. Journal of Marketing  \textbf{81}(1),  36--58 (2017)

\bibitem{palmirani2018legal}
Palmirani, M., Martoni, M., Rossi, A., Bartolini, C., Robaldo, L.: Legal
  ontology for modelling gdpr concepts and norms. In: JURIX. pp. 91--100 (2018)

\bibitem{ravichander2019question}
Ravichander, A., Black, A.W., Wilson, S., Norton, T., Sadeh, N.: Question
  answering for privacy policies: Combining computational and legal
  perspectives. arXiv preprint arXiv:1911.00841  (2019)

\bibitem{reidenberg2016ambiguity}
Reidenberg, J.R., Bhatia, J., Breaux, T.D., Norton, T.B.: Ambiguity in privacy
  policies and the impact of regulation. The Journal of Legal Studies
  \textbf{45}(S2),  S163--S190 (2016)

\bibitem{sadeh2013usable}
Sadeh, N., Acquisti, A., Breaux, T.D., Cranor, L.F., McDonald, A.M.,
  Reidenberg, J.R., Smith, N.A., Liu, F., Russell, N.C., Schaub, F., et~al.:
  The usable privacy policy project: Combining crowdsourcing. Machine Learning
  and Natural Language Processing to Semi-Automatically Answer Those Privacy
  Questions Users Care About. Carnegie Mellon University Technical Report
  CMU-ISR-13-119 pp. 1--24 (2013)

\bibitem{shvartzshnaider2018analyzing}
Shvartzshnaider, Y., Apthorpe, N., Feamster, N., Nissenbaum, H.: Analyzing
  privacy policies using contextual integrity annotations. arXiv preprint
  arXiv:1809.02236  (2018)

\bibitem{tesfay2018read}
Tesfay, W.B., Hofmann, P., Nakamura, T., Kiyomoto, S., Serna, J.: I read but
  don't agree: Privacy policy benchmarking using machine learning and the eu
  gdpr. In: Companion Proceedings of the The Web Conference 2018. pp. 163--166.
  International World Wide Web Conferences Steering Committee (2018)

\bibitem{stourm2019}
V.~Stourm, D. Restrepo~Amariles, S.N.: A framework for managers to evaluate and
  respond to privacy regulation  (2019)

\bibitem{watanabe2015understanding}
Watanabe, T., Akiyama, M., Sakai, T., Mori, T.: Understanding the
  inconsistencies between text descriptions and the use of privacy-sensitive
  resources of mobile apps. In: Eleventh Symposium On Usable Privacy and
  Security ($\{$SOUPS$\}$ 2015). pp. 241--255 (2015)

\bibitem{wilson2016creation}
Wilson, S., Schaub, F., Dara, A.A., Liu, F., Cherivirala, S., Leon, P.G.,
  Andersen, M.S., Zimmeck, S., Sathyendra, K.M., Russell, N.C., et~al.: The
  creation and analysis of a website privacy policy corpus. In: Proceedings of
  the 54th Annual Meeting of the Association for Computational Linguistics
  (Volume 1: Long Papers). pp. 1330--1340 (2016)

\bibitem{wilson2018analyzing}
Wilson, S., Schaub, F., Liu, F., Sathyendra, K.M., Smullen, D., Zimmeck, S.,
  Ramanath, R., Story, P., Liu, F., Sadeh, N., et~al.: Analyzing privacy
  policies at scale: From crowdsourcing to automated annotations. ACM
  Transactions on the Web (TWEB)  \textbf{13}(1), ~1 (2018)

\bibitem{zimmeck2014privee}
Zimmeck, S., Bellovin, S.M.: Privee: An architecture for automatically
  analyzing web privacy policies. In: 23rd $\{$USENIX$\}$ Security Symposium
  ($\{$USENIX$\}$ Security 14). pp. 1--16 (2014)

\bibitem{zimmeck2019maps}
Zimmeck, S., Story, P., Smullen, D., Ravichander, A., Wang, Z., Reidenberg, J.,
  Russell, N.C., Sadeh, N.: Maps: Scaling privacy compliance analysis to a
  million apps. Proceedings on Privacy Enhancing Technologies
  \textbf{2019}(3),  66--86 (2019)

\end{thebibliography}

\end{document}